\documentclass[10pt,twocolumn,letterpaper]{article}

\usepackage[pagenumbers]{cvpr} 

\usepackage{booktabs}  
\usepackage{tabularx}  
\usepackage{array}     
\usepackage{multirow}  
\usepackage{xcolor}    
\usepackage[table]{xcolor}
\definecolor{cvprblue}{rgb}{0.21,0.49,0.74}
\usepackage[pagebackref,breaklinks,colorlinks,allcolors=cvprblue]{hyperref}

\def\confName{CVPR}
\def\confYear{2027}

\title{Support-Guided Triadic Decoding for Hallucination Mitigation \\ in Multimodal Dialogue with LVLMs}

\makeatletter
\newcommand{\myfnsymbol}[1]{%
  \expandafter\@myfnsymbol\csname c@#1\endcsname
}
\newcommand{\@myfnsymbol}[1]{%
  \ifcase #1
  \or \TextOrMath{\textasteriskcentered}{*}
  \or \TextOrMath{\textdagger}{\dagger}
  \fi
}
\newcommand{\equalcontributor}{\@myfnsymbol{1}}
\newcommand{\corresponding}{\@myfnsymbol{2}}
\makeatother

\author{
Yiheng Li$^{1,2}$,
Wenhao Wang$^{3,\corresponding}$,
Xingyi Li$^{4}$,
Zhilin Tang$^{1,2}$,\\
Zichang Tan$^{6}$,
Zhen Lei$^{1,2,5,7}$,
Yang Yang$^{1,2,\corresponding}$\\
$^1$ School of Artificial Intelligence, University of Chinese Academy of Sciences\\
$^2$ MAIS, Institute of Automation, Chinese Academy of Sciences\\
$^3$ Vast Intelligence Lab,
$^4$ Institute of Information Engineering, Chinese Academy of Sciences\\
$^5$ CAIR, HKISI, Chinese Academy of Sciences, $^6$ Sangfor Technologies Inc.
$^7$ SCSE, FIE, M.U.S.T\\
{\tt\small \{liyiheng2024, yangyang2013, zhen.lei\}@ia.ac.cn, wangwenhao@vastilab.com} 
}

\begin{document}
\maketitle
\renewcommand{\thefootnote}{\myfnsymbol{footnote}}
\footnotetext[2]{Corresponding authors}%
\begin{abstract}
In multimodal multi-turn dialogue, dialogue history plays a dual role: reliable responses provide useful context for subsequent turns, whereas hallucinated responses can propagate errors across turns. Existing training-free methods globally adjust the effect of dialogue history on generation, making it difficult to suppress misleading information without weakening useful context. To address this trade-off, we introduce Support-Guided Triadic Decoding (SGTD), a training-free framework that selectively controls the influence of unreliable history and fuses complementary predictions during decoding. Specifically, Historical Answer Support Estimation measures the visual support of historical responses against type-matched alternatives and identifies low-support content that may mislead later generation. Support-Targeted Attention Intervention then weakens the influence of this content while preserving the surrounding dialogue, producing a support-refined prediction. Triadic Decoding combines this prediction with the full-history prediction and answer-free visual evidence, while the Support-Adaptive Plausibility Constraint filters implausible tokens using visual agreement and historical support. Experiments on MMHalSnowball and VisDiaHalBench across representative LVLMs show that SGTD consistently achieves the highest accuracy under hallucinated histories while preserving the benefits of factual context.
\end{abstract}    
\section{Introduction}
\label{sec:introduction}

\begin{figure}[t]
    \centering
    \includegraphics[width=\columnwidth]{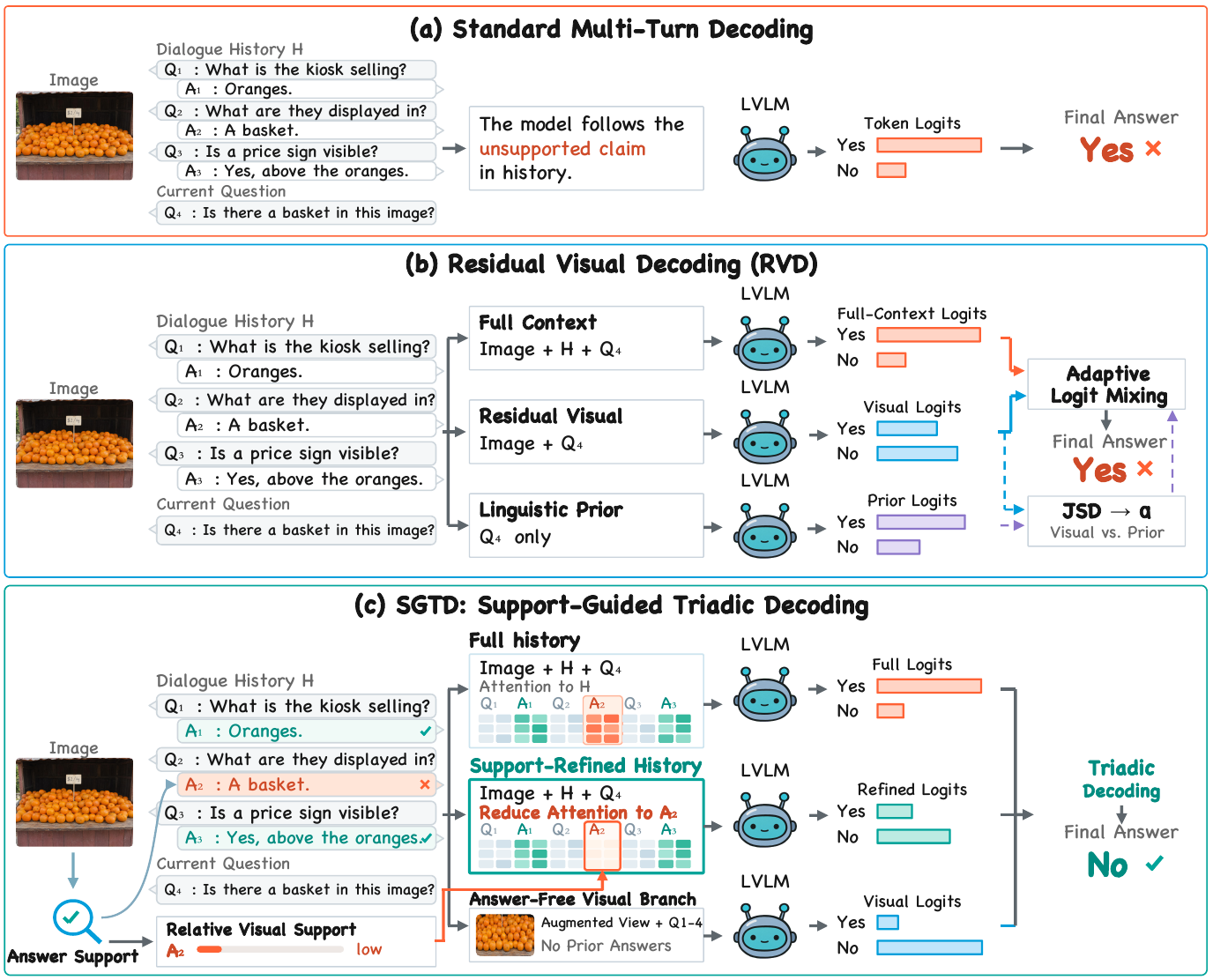}
    \caption{\textbf{Comparison of decoding strategies under unreliable dialogue history.} (a) Standard decoding can follow an unsupported prior answer. (b) RVD applies a distribution-level correction without localizing unreliable history, potentially weakening reliable context. (c) SGTD uses answer support to refine low-support history, then jointly decodes from the Full-History, Support-Refined History, and Answer-Free Visual Branches.}
    \label{fig:sgtd_overview}
\end{figure}

In multimodal multi-turn settings, previous dialogue helps large vision-language models (LVLMs) resolve follow-up questions and maintain coherence~\cite{das2017visualdialog,lee2025multiverse}.
Their value, however, depends on the reliability of the accumulated context.
As shown in Fig.~\ref{fig:sgtd_overview} (a), a hallucinated answer carried into the dialogue history can override contradictory visual evidence and distort a later response~\cite{park2024mitigating,cao2024visdiahalbench,zhong2024investigating}.
A reliable dialogue system should therefore place greater weight on well-supported history while reducing the influence of unreliable content on subsequent generation.

Existing methods improve visual faithfulness through robust instruction tuning~\cite{liu2024robust}, preference alignment~\cite{yu2024rlhfv,wang2024mdpo}, or training-free inference.
Among these, training-free approaches require no data or parameter updates, making them suitable for deployment.
They use contrastive decoding~\cite{leng2024vcd,chen2024halc,huo2025sid}, attention-guided decoding~\cite{huang2024opera}, or attention intervention~\cite{liu2024pai,wan2025only,li2026cmac}.
Yet most target single-turn generation and do not account for an evolving dialogue history.
RVD~\cite{zhong2024investigating} uses Jensen--Shannon divergence between visual and question-only distributions to weight full-context and history-free visual logits, as shown in Fig.~\ref{fig:sgtd_overview} (b).
CAVR~\cite{jiang2026mmsnowball} uses layer-wise uncertainty to refresh visual representations, then rectifies output logits by blending full-context and history-free visual predictions.
Despite intervening at different stages, both use signals to correct history influence as a whole, leaving reliable and unreliable content undifferentiated.

Within the decoder, the attention mechanism carries information from earlier dialogue turns into the hidden states used for next-token prediction.
Unsupported history can therefore bias these representations before the next-token logits are produced.
Output-level uncertainty or distributional corrections may reshape the prediction, but do not isolate the contribution of unreliable context within these representations.
A stronger global correction~\cite{zhong2024investigating,jiang2026mmsnowball} may reduce this bias, but it can also weaken useful information carried across turns.
Effective mitigation should instead regulate the influence of unreliable history within the decoder while preserving well-supported context.

Visual support estimation offers a way to connect content reliability with the history spans that shape later representations.
Fine-grained factuality methods assess responses at the claim or token level~\cite{min2023factscore,jing2024faithscore,park2025halloc}, while post-hoc systems such as Woodpecker~\cite{yin2024woodpecker} and Pelican~\cite{sahu2024pelican} check decomposed claims against the image.
In multi-turn dialogue, identifying low-support content is only the first step.
Controlling its influence while preserving useful context calls for three aligned predictions.
A full-history prediction retains the complete dialogue, a support-refined prediction reflects the effect of weakening low-support spans, and an answer-free visual prediction provides grounding independent of previous answers.
As illustrated in Fig.~\ref{fig:sgtd_overview} (c), combining these predictions keeps context preservation, targeted refinement, and answer-independent visual grounding distinct rather than forcing one globally corrected prediction to serve all three roles.

Based on this insight, we introduce Support-Guided Triadic Decoding (SGTD), a training-free framework that uses turn-level visual support to coordinate the three predictions during decoding.
Historical Answer Support Estimation first resolves cross-turn references and uses a frozen VQA scorer~\cite{li2022blip} to compare each checkable historical answer with type-matched alternatives.
The resulting relative support scores select low-support turns, whose answers are mapped to token spans in the dialogue history.
Support-Targeted Attention Intervention then reduces attention to these selected spans while retaining the surrounding dialogue, yielding the support-refined prediction.
The full-history prediction provides a reference, while the Answer-Free Visual Branch combines previous questions with the most relevant regional image view.
Triadic Decoding combines the support-refined and answer-free visual predictions, then adjusts the result using the difference between the support-refined and full-history predictions.
It then filters unlikely candidates according to how well the current prediction agrees with the image and how reliable the history appears.
The final prediction relies less on unreliable history while retaining useful context across turns.

We evaluate SGTD on MMHalSnowball~\cite{zhong2024investigating} and VisDiaHalBench~\cite{cao2024visdiahalbench} across representative LVLMs.
SGTD consistently achieves the highest accuracy under hallucinated histories and the best mean performance while retaining the gains provided by factual history.
Compared with RVD, SGTD improves robustness to misleading context while achieving higher factual-history accuracy.
It also yields the lowest Hallucination Drop across both benchmarks, indicating more stable predictions when hallucinated history is introduced.
These results show that SGTD balances the suppression of misleading history with the preservation of useful cross-turn context. Our contributions are threefold:
\begin{itemize}
    \item We frame robust multimodal dialogue decoding as preserving useful history while separating the predictive effect of low-support content.
    \item We introduce SGTD, a training-free framework that combines Historical Answer Support Estimation, Support-Targeted Attention Intervention, and Triadic Decoding.
    \item We evaluate SGTD on two multimodal dialogue benchmarks with representative LVLMs, consistently achieving the best accuracy under hallucinated histories while retaining gains from factual context.
\end{itemize}

\section{Related Work}
\label{sec:related}

\subsection{Hallucination in Multimodal Dialogue}
Multimodal dialogue extends image-grounded interaction across turns, where history supports follow-up reasoning and visual coreference~\cite{das2017visualdialog,kottur2018coreference} but may also encourage contextual shortcuts over visual evidence~\cite{qi2020causal}.
ConvBench~\cite{liu2024convbench} studies hierarchical dialogue, MMDU~\cite{liu2024mmdu} evaluates long multi-image conversations, and MultiVerse~\cite{lee2025multiverse} compares oracle with self-generated context to expose errors from prior answers.
Hallucination-oriented benchmarks examine unreliable history directly.
EvalDial~\cite{park2024mitigating} uses adversarial dialogue to measure textual-context bias, while VisDiaHalBench~\cite{cao2024visdiahalbench} tests misleading histories with cross-turn references.
MMHalSnowball~\cite{zhong2024investigating} isolates cases where a hallucinated answer changes a later prediction, while MM-Snowball~\cite{jiang2026mmsnowball} studies how such errors accumulate across turns.
Together, these benchmarks reveal unreliable history as a distinct source of hallucination that can distort later responses and compound across turns.

\begin{figure*}[t]
    \centering
    \includegraphics[width=\textwidth]{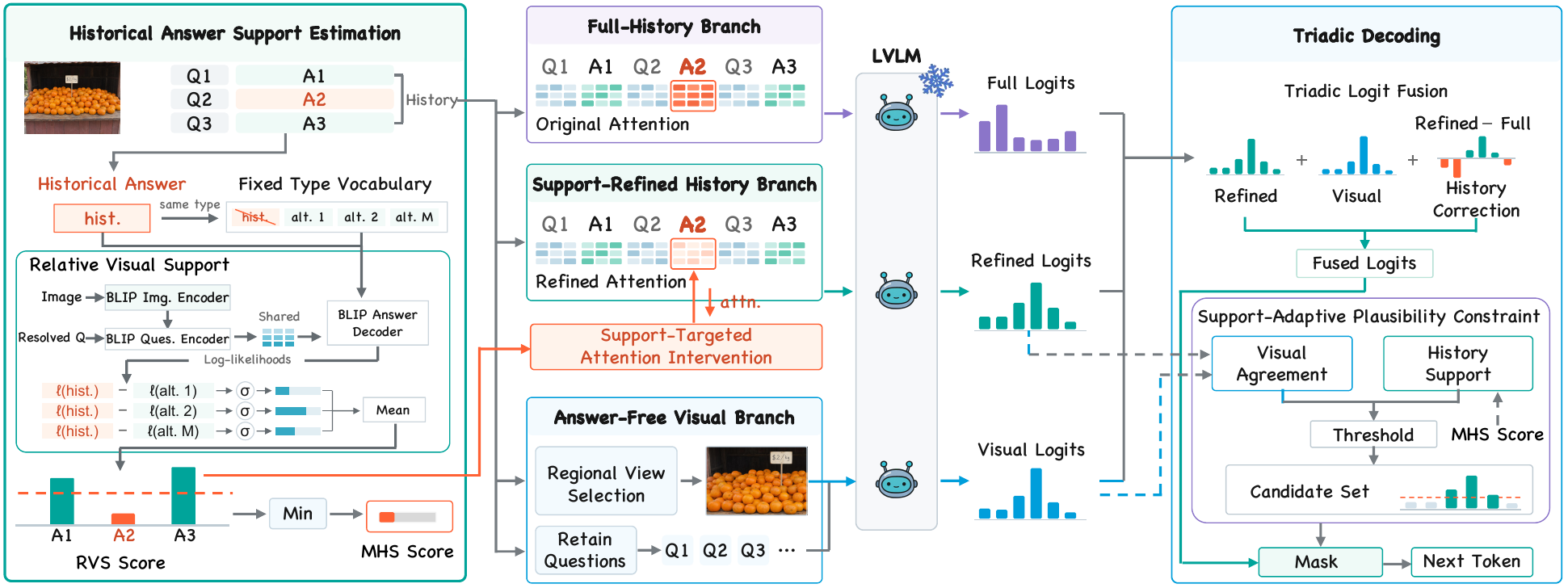}
    \caption{\textbf{Overview of Support-Guided Triadic Decoding (SGTD).} Frozen BLIP compares checkable historical answers with type-matched alternatives to identify low-support turns. The Full-History, Support-Refined History, and Answer-Free Visual Branches preserve context, reduce attention to selected answer spans, and provide visual evidence independent of previous answers. Triadic Decoding combines their logits under the Support-Adaptive Plausibility Constraint.}
    \label{fig:sgtd_method_overview}
\end{figure*}

\subsection{Training-Free Hallucination Mitigation}
Training-free methods mitigate hallucination without task-specific parameter optimization by reshaping output distributions or modifying internal computation.
Early distribution-level methods construct auxiliary predictions by weakening or perturbing conditioning signals.
VCD~\cite{leng2024vcd} contrasts predictions from original and distorted images, M3ID~\cite{favero2024m3id} contrasts conditional and unconditional predictions, and ICD~\cite{wang2024icd} uses disturbance instructions.
OPERA~\cite{huang2024opera} penalizes over-trust in summary tokens during decoding.
Later methods make the auxiliary signal more targeted.
AGLA~\cite{an2025agla} combines global perception with a query-local view, while SID~\cite{huo2025sid} constructs a self-introspective branch from visual tokens retained after the early decoder layers.
Internal interventions instead modify representations or cached states before token selection.
ClearSight~\cite{yin2025clearsight} and VISTA~\cite{li2025vista} strengthen visual information in intermediate layers.
ONLY~\cite{wan2025only} intervenes in a single decoder layer, Nullu~\cite{yang2025nullu} projects out a learned hallucination subspace, and PTI~\cite{zhang2026pti} steers the prefill cache.
Most of these methods assume a fixed context and target single-turn generation.
For multi-turn dialogue, RVD~\cite{zhong2024investigating} adds a residual visual correction to the full-history prediction, while CAVR~\cite{jiang2026mmsnowball} combines uncertainty-triggered visual refreshing with logit rectification.
Both rely on global signals rather than localizing unreliable history.
SGTD instead uses turn-level support to guide selective attention intervention and Triadic Decoding.

\subsection{Fine-Grained Factuality Verification}
Fine-grained factuality verification evaluates individual facts or spans rather than assigning one judgment to an entire response.
FActScore~\cite{min2023factscore} decomposes long-form text into atomic facts and checks them against external evidence, while FaithScore~\cite{jing2024faithscore} verifies descriptive facts in multimodal responses against the image.
Span-level methods further localize unsupported content.
HalLoc~\cite{park2025halloc} provides probabilistic token-level annotations and detection, while MHALO~\cite{cai2025mhalo} evaluates hallucination span identification and categorization.
Verification can also support post-hoc correction.
Woodpecker~\cite{yin2024woodpecker} builds a visual knowledge base from formulated questions before rewriting inconsistent content, whereas Pelican~\cite{sahu2024pelican} verifies predicate--question subclaims through tool composition.
These methods verify or revise the response under inspection, whereas SGTD uses turn-level support to regulate how earlier responses influence later decoding.

\section{Method}
\label{sec:method}

\subsection{Problem Formulation}
\label{sec:problem_formulation}

Consider an image-grounded dialogue at turn $t$.
The model conditions on an image $I$, the current question $q_t$, and history $\mathcal{H}_t=((q_i,a_i))_{i=1}^{t-1}$, where $(q_i,a_i)$ is the question--answer pair from turn $i$.
A frozen LVLM $f_{\theta}$ autoregressively generates the current answer $a_t$.
At generation step $k$, the model produces the next-token logits:
\begin{equation}
    \mathbf{z}_{t,k}
    =f_{\theta}\!\left(I,\mathcal{H}_t,q_t,a_{t,<k}\right)
    \in\mathbb{R}^{|\mathcal{V}|},
    \label{eq:full_history_logits}
\end{equation}
where $a_{t,<k}=(a_{t,1},\ldots,a_{t,k-1})$ denotes the generated prefix and $\mathcal{V}$ is the output vocabulary.
We evaluate generation under three history conditions: no dialogue history ($\mathcal{H}_t=\varnothing$), a history with factual responses, and a history with hallucinated responses, referred to as Clean, Fact., and Hallu., respectively.

\subsection{Overview}
\label{sec:method_overview}

Support-Guided Triadic Decoding (SGTD) follows the inference route in Fig.~\ref{fig:sgtd_method_overview}.
Historical Answer Support Estimation compares each checkable historical answer with type-matched alternatives to identify low-support turns and locate their answer tokens.
Support-Targeted Attention Intervention reduces attention to these tokens to form the Support-Refined History Branch, while the Full-History Branch provides the reference prediction.
The Answer-Free Visual Branch retains earlier questions, removes previous answers, and selects the regional image view most relevant to the current question.
Triadic Decoding combines the support-refined and answer-free visual logits and adds the difference between the support-refined and full-history logits as a targeted correction.
The Support-Adaptive Plausibility Constraint then filters the result using visual agreement and minimum historical support.

\subsection{Historical Answer Support Estimation}
\label{sec:history_support_estimation}

The full-history logits $\mathbf{z}_{t,k}$ aggregate information from all preceding turns but do not reveal which parts of the context conflict with the image.
SGTD therefore assigns a Relative Visual Support score to each checkable historical answer and constructs a declarative claim for token localization.
The scores select low-support turns, whose answer spans enter the Support-Refined History Branch.

\noindent\textbf{Answer-Set Construction.}
For each turn $i<t$, we replace references in $q_i$ with the corresponding entities from earlier dialogue turns and denote the resulting standalone question by $\bar q_i$.
The historical answer $a_i$ serves as the candidate $u_i$.
We form the answer set $\mathcal{U}_i=\{u_i\}\cup\mathcal{U}_i^{-}$, where $\mathcal{U}_i^{-}=\{u_{i,m}^{-}\}_{m=1}^{M_i}$ contains $M_i$ type-matched alternatives.
Every $u\in\mathcal{U}_i$ is paired with $\bar q_i$, so only the proposed answer varies.
For binary questions, $\mathcal{U}_i^{-}$ contains the opposite answer.
For color, count, and spatial relation questions, it is drawn from a fixed vocabulary of the same type after excluding $u_i$.
The construction uses answer-type templates and adds negligible preprocessing overhead.
Unanswerable or ambiguous turns and those without a valid alternative set are left unscored.
The pair $(\bar q_i,\mathcal{U}_i)$ is passed to the support estimator.
We then form a claim $c_i$ by turning $\bar q_i$ into a statement and completing it with $a_i$.
The claim is later used by Support-Targeted Attention Intervention.

\noindent\textbf{Relative Visual Support.}
Given $(\bar q_i,\mathcal{U}_i)$, frozen BLIP~\cite{li2022blip} encodes $(I,\bar q_i)$ once and assigns each $u\in\mathcal{U}_i$ a log-likelihood $\ell_i(u)$.
We compare $u_i$ with every $u_{i,m}^{-}$ rather than using its likelihood alone.
Each score difference is converted into a pairwise preference, whose mean defines the Relative Visual Support (RVS) score $s_i$:
\begin{equation}
    s_i=\frac{1}{M_i}\sum_{m=1}^{M_i}
    \sigma\!\left(\ell_i(u_i)-\ell_i(u_{i,m}^{-})\right),
    \label{eq:relative_visual_support}
\end{equation}
where $\sigma$ is the sigmoid function.
Each term gives the pairwise support of the historical answer $u_i$ against one type-matched alternative.
The average keeps $s_i\in[0,1]$ on a consistent scale regardless of $M_i$.
When $s_i>0.5$, $u_i$ receives greater pairwise support on average.
Lower values indicate stronger visual support for the alternatives in $\mathcal{U}_i^{-}$.

\noindent\textbf{Low-Support Selection.}
We select low-support turns by comparing their RVS scores with a threshold $\tau_{\mathrm{sup}}\in(0,1)$:
\begin{equation}
    \mathcal{L}_t=\{i<t\mid s_i\leq\tau_{\mathrm{sup}}\}.
    \label{eq:low_support_turns}
\end{equation}
$\mathcal{L}_t$ determines which answer spans enter the attention intervention.
The smallest RVS score among the scored history turns defines the minimum historical support (MHS) score $s_t^{\mathrm{hist}}$, which summarizes the weakest visual support available to the Support-Adaptive Plausibility Constraint.
Unscored dialogue turns enter neither the MHS score nor the attention intervention.
If the dialogue has no history or all history turns are unscored, we set the MHS score to the neutral value $s_t^{\mathrm{hist}}=1$.

\subsection{Triadic Branch Construction}
\label{sec:triadic_branch_construction}

Using the estimated support, SGTD evaluates each next token with three aligned forward passes through the frozen LVLM.
All branches share the generated prefix, but differ in how they use previous answers and visual input.
The three predictions preserve the complete history, expose the effect of targeted attention intervention, and provide visual evidence independent of previous answers.

\noindent\textbf{Full-History Branch.}
The Full-History Branch follows standard decoding with the original image and complete dialogue history.
It produces the reference logits $\mathbf{z}_{t,k}$ in Eq.~\ref{eq:full_history_logits}.

\begin{figure}[t]
    \centering
    \includegraphics[width=\columnwidth]{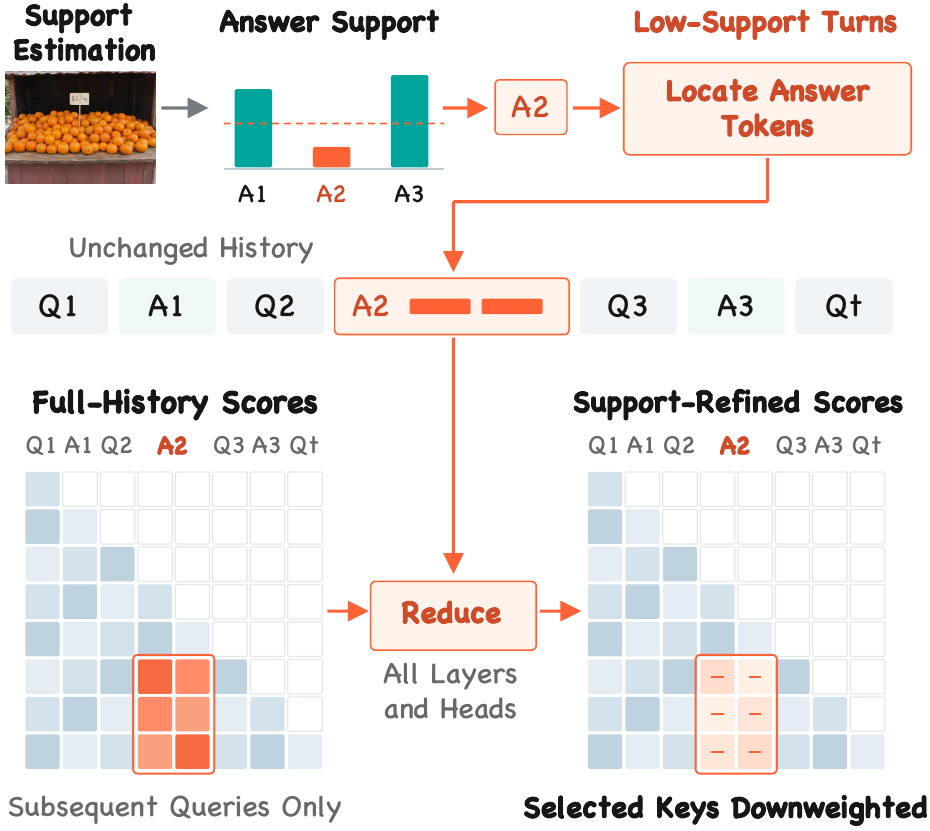}
    \caption{\textbf{Support-Targeted Attention Intervention.} Declarative claims from low-support turns locate the corresponding answer tokens. SGTD subtracts a penalty from the pre-softmax attention scores directed to these tokens by subsequent queries.}
    \label{fig:sgtd_attention_intervention}
\end{figure}

\noindent\textbf{Support-Refined History Branch.}
The Support-Refined History Branch retains the complete dialogue while selectively reducing the influence of low-support content.
For each $i\in\mathcal{L}_t$, the answer content represented by $u_i$ and $c_i$ is mapped to its token positions in the expanded multimodal sequence, denoted by $\mathcal{S}_i$.
As illustrated in Fig.~\ref{fig:sgtd_attention_intervention}, \textit{Support-Targeted Attention Intervention} acts only on $\mathcal{S}_i$, preserving the remaining answer content and surrounding dialogue.

Let $A_{\ell,h}(x,y)$ denote the pre-softmax attention score from query position $x$ to key position $y$ in decoder layer $\ell$ and attention head $h$.
SGTD subtracts a penalty $\lambda>0$ from scores directed to each selected span:
\begin{equation}
    \widetilde{A}_{\ell,h}(x,y)
    =A_{\ell,h}(x,y)-\lambda
    \sum_{i\in\mathcal{L}_t}
    \mathbf{1}[y\in\mathcal{S}_i]\mathbf{1}[x>\max(\mathcal{S}_i)].
    \label{eq:history_attention_penalty}
\end{equation}
$\mathbf{1}[\cdot]$ is the indicator function, and $x>\max(\mathcal{S}_i)$ restricts the penalty to queries after the selected answer.
Within this branch, the support score determines span selection but not the penalty magnitude.
The penalty $\lambda$ is applied uniformly to every selected span across all decoder layers and heads.
After softmax, the reduced scores lower the contribution of these spans to subsequent token representations.
The modified scores $\widetilde{A}$ produce the support-refined history logits:
\begin{equation}
    \mathbf{z}_{t,k}^{\mathrm{ref}}
    =f_{\theta}\!\left(
    I,\mathcal{H}_t,q_t,a_{t,<k};\widetilde{A}
    \right).
    \label{eq:support_refined_logits}
\end{equation}
When $\mathcal{L}_t=\varnothing$, the intervention is inactive and $\mathbf{z}_{t,k}^{\mathrm{ref}}=\mathbf{z}_{t,k}$.

\noindent\textbf{Answer-Free Visual Branch.}
The Answer-Free Visual Branch complements the two history branches with a visual prediction centered on the current question.
To focus the visual evidence spatially, we derive five overlapping regional image views $\{I_j\}_{j=1}^{5}$ that respectively cover the four image corners and the center of $I$.
Each view retains its target region and softly suppresses the surrounding image content.
A frozen image--text matching (ITM) model~\cite{li2022blip} evaluates the agreement between each view and the current question $q_t$.
We select the most relevant view as $\widehat{I}_t=I_{j^\star}$, where $j^\star=\arg\max_j \operatorname{ITM}(I_j,q_t)$.
For its dialogue context, the branch uses $\mathcal{Q}_t=(q_1,\ldots,q_t)$, which retains the ordered questions through turn $t$ and excludes all previous answers.
The resulting answer-free visual logits are:
\begin{equation}
    \mathbf{z}_{t,k}^{\mathrm{vis}}
    =f_{\theta}\!\left(
    \widehat{I}_t,\mathcal{Q}_t,a_{t,<k}
    \right).
    \label{eq:answer_free_visual_logits}
\end{equation}
These logits provide visual evidence for logit fusion and are also used to compute the visual agreement in the Support-Adaptive Plausibility Constraint.

\subsection{Triadic Decoding}
\label{sec:triadic_decoding}

Triadic Decoding combines the three branch logits into a prediction.
It limits sampling to candidates supported by the Support-Refined History Branch, with the filtering strength adapted to visual agreement and historical support.

\noindent\textbf{Triadic Logit Fusion.}
Attention intervention changes the support-refined history logits, but these logits alone do not reveal how the prediction has shifted.
The full-history logits provide a matched reference for measuring this shift.
Their difference $\mathbf{z}_{t,k}^{\mathrm{ref}}-\mathbf{z}_{t,k}$ therefore isolates the score change caused by reducing attention to the selected history spans.
SGTD first combines the support-refined and answer-free visual logits, then adds this difference as a targeted history correction.
Following the paired prediction formulation in contrastive decoding~\cite{li2023contrastive}, the fused logits are:
\begin{equation}
    \widetilde{\mathbf{z}}_{t,k}
    =\mathbf{z}_{t,k}^{\mathrm{ref}}
    +\omega\mathbf{z}_{t,k}^{\mathrm{vis}}
    +(1+\omega)\alpha_{\mathrm{hist}}
    \left(\mathbf{z}_{t,k}^{\mathrm{ref}}-\mathbf{z}_{t,k}\right).
    \label{eq:triadic_fusion}
\end{equation}
Here, $\omega\geq0$ controls the contribution of answer-free visual evidence, while $\alpha_{\mathrm{hist}}\geq0$ sets the strength of the residual correction.
The first two terms form a base prediction that combines support-refined dialogue context with regional visual evidence selected for $q_t$.
The residual term favors tokens whose scores rise after the intervention and suppresses those whose scores fall because they rely on the selected spans.
The factor $(1+\omega)$ keeps this correction proportional to the combined weight of the two base terms.

\noindent\textbf{Support-Adaptive Plausibility Constraint.}
Logit fusion can elevate candidates that remain weak under the support-refined history prediction.
SGTD therefore applies plausibility filtering~\cite{li2023contrastive,leng2024vcd,an2025agla}, anchoring the candidate set to $\mathbf{z}_{t,k}^{\mathrm{ref}}$ and adapting its selectivity to visual agreement and the MHS score.
Let $\widehat{w}_{t,k}^{\mathrm{ref}}=\arg\max_{w\in\mathcal{V}}\mathbf{z}_{t,k}^{\mathrm{ref}}(w)$ be the top token in $\mathbf{z}_{t,k}^{\mathrm{ref}}$.
The visual agreement score for this token is:
\begin{equation}
    g_{t,k}=\exp\!\left(
    \mathbf{z}_{t,k}^{\mathrm{vis}}(\widehat{w}_{t,k}^{\mathrm{ref}})
    -\max_{w\in\mathcal{V}}\mathbf{z}_{t,k}^{\mathrm{vis}}(w)
    \right),
    \label{eq:visual_agreement}
\end{equation}
where $g_{t,k}\in(0,1]$ equals $1$ when the token also has the largest visual logit and decreases as the gap from the visual maximum grows.
Together with the MHS score, this agreement sets the adaptive threshold:
\begin{equation}
    \beta_{t,k}=\beta_{\min}
    +g_{t,k}s_t^{\mathrm{hist}}\left(\beta_{\max}-\beta_{\min}\right),
    \label{eq:adaptive_beta}
\end{equation}
where $0<\beta_{\min}\leq\beta_{\max}\leq1$ bound its range.
When both signals are large, $\beta_{t,k}$ moves toward $\beta_{\max}$, bringing the cutoff closer to the maximum support-refined logit and narrowing the candidate set.
If either signal is weak, $\beta_{t,k}$ approaches $\beta_{\min}$ and admits a broader set.
A token $w$ remains eligible only if $\mathbf{z}_{t,k}^{\mathrm{ref}}(w)\geq\max_{v\in\mathcal{V}}\mathbf{z}_{t,k}^{\mathrm{ref}}(v)+\log\beta_{t,k}$.
Eligible tokens keep their scores in $\widetilde{\mathbf{z}}_{t,k}$, while all others are masked before sampling.

\noindent\textbf{History-Free Decoding.}
When $\mathcal{H}_t=\varnothing$, corresponding to the Clean setting, no historical answers are available for support estimation and no answer spans require attention intervention.
The Full-History Branch and Support-Refined History Branch therefore coincide, giving $\mathbf{z}_{t,k}^{\mathrm{ref}}=\mathbf{z}_{t,k}$ and removing the residual history correction from Eq.~\ref{eq:triadic_fusion}.
The Answer-Free Visual Branch, meanwhile, remains active with $\mathcal{Q}_t=(q_t)$, so the fused logits reduce to $\widetilde{\mathbf{z}}_{t,k}=\mathbf{z}_{t,k}+\omega\mathbf{z}_{t,k}^{\mathrm{vis}}$.
The neutral MHS score $s_t^{\mathrm{hist}}=1$ similarly reduces Eq.~\ref{eq:adaptive_beta} to $\beta_{t,k}=\beta_{\min}+g_{t,k}(\beta_{\max}-\beta_{\min})$.
As a result, Clean decoding retains regional visual guidance, while visual agreement alone determines the selectivity of plausibility filtering.

\section{Experiments}
\label{sec:experiments}

\begin{table*}[t]
\centering

\caption{\textbf{Results on MMHalSnowball (\%).} Avg. is the mean over Clean, Fact., and Hallu., while H.D. is Clean minus Hallu. Under the Clean setting, RVD reduces to Regular decoding. Best and second-best results are shown in bold and underlined.}
\label{tab:mmhalsnowball_baselines}

\begingroup
\scriptsize
\setlength{\tabcolsep}{1.3pt}
\renewcommand{\arraystretch}{0.96}

\begin{tabularx}{\linewidth}{
    >{\raggedright\arraybackslash}p{1.55cm}
    *{5}{>{\centering\arraybackslash}X}
    @{\hspace{1.5pt}}
    !{\color{black!25}\vrule width 0.45pt}
    @{\hspace{1.5pt}}
    *{5}{>{\centering\arraybackslash}X}
    @{\hspace{1.5pt}}
    !{\color{black!25}\vrule width 0.45pt}
    @{\hspace{1.5pt}}
    *{5}{>{\centering\arraybackslash}X}
}

\toprule
\multirow{2}{*}{\textbf{Method}}
& \multicolumn{5}{c}{\textbf{LLaVA-1.5-7B}}
& \multicolumn{5}{c}{\textbf{mPLUG-Owl2-7B}}
& \multicolumn{5}{c}{\textbf{ShareGPT4V-7B}}
\tabularnewline

\cmidrule(lr){2-6}
\cmidrule(lr){7-11}
\cmidrule(lr){12-16}

& \mbox{Clean $\uparrow$}
& \mbox{Fact. $\uparrow$}
& \mbox{Hallu. $\uparrow$}
& \mbox{Avg. $\uparrow$}
& \mbox{H.D. $\downarrow$}
& \mbox{Clean $\uparrow$}
& \mbox{Fact. $\uparrow$}
& \mbox{Hallu. $\uparrow$}
& \mbox{Avg. $\uparrow$}
& \mbox{H.D. $\downarrow$}
& \mbox{Clean $\uparrow$}
& \mbox{Fact. $\uparrow$}
& \mbox{Hallu. $\uparrow$}
& \mbox{Avg. $\uparrow$}
& \mbox{H.D. $\downarrow$}
\tabularnewline

\midrule

Regular
& $71.00$
& $88.74$
& $14.72$
& $58.15$
& $56.28$
& $60.67$
& $87.69$
& $7.76$
& $52.04$
& $52.91$
& $73.56$
& $90.63$
& $15.60$
& $59.93$
& $57.95$
\tabularnewline

VCD~\cite{leng2024vcd}
& $70.02$
& $88.84$
& $17.60$
& $58.82$
& $52.42$
& $61.51$
& $88.62$
& $8.57$
& $52.90$
& $52.95$
& $73.56$
& $90.63$
& $19.18$
& $61.12$
& $54.37$
\tabularnewline

M3ID~\cite{favero2024m3id}
& $72.37$
& $\underline{90.27}$
& $15.02$
& $59.22$
& $57.35$
& $62.54$
& $\underline{88.88}$
& $7.22$
& $52.88$
& $55.32$
& $74.40$
& $91.76$
& $15.79$
& $60.65$
& $58.62$
\tabularnewline

PAI~\cite{liu2024pai}
& $72.83$
& $89.66$
& $15.95$
& $59.48$
& $56.89$
& $62.50$
& $88.76$
& $7.76$
& $53.01$
& $54.74$
& $74.72$
& $91.59$
& $16.91$
& $61.08$
& $57.81$
\tabularnewline

SID~\cite{huo2025sid}
& $72.91$
& $89.42$
& $20.61$
& $60.98$
& $52.30$
& $60.18$
& $87.89$
& $8.91$
& $52.33$
& $51.28$
& $75.25$
& $90.35$
& $21.34$
& $62.31$
& $53.91$
\tabularnewline

AvisC~\cite{woo2025avisc}
& $75.11$
& $89.99$
& $20.03$
& $61.71$
& $55.08$
& $61.17$
& $85.46$
& $11.68$
& $52.77$
& $49.49$
& $\underline{76.79}$
& $\mathbf{92.32}$
& $17.53$
& $62.21$
& $59.26$
\tabularnewline

CAI~\cite{lei2026cai}
& $\underline{75.31}$
& $85.42$
& $26.64$
& $62.46$
& $48.66$
& $\underline{63.34}$
& $86.19$
& $11.62$
& $53.72$
& $51.72$
& $76.11$
& $87.75$
& $26.24$
& $63.37$
& $49.87$
\tabularnewline

\midrule

RVD~\cite{zhong2024investigating}
& $71.00$
& $83.87$
& $\underline{32.62}$
& $\underline{62.50}$
& $\underline{38.39}$
& $60.67$
& $81.82$
& $\underline{22.14}$
& $\underline{54.88}$
& $\underline{38.53}$
& $73.56$
& $85.84$
& $\underline{33.84}$
& $\underline{64.41}$
& $\underline{39.71}$
\tabularnewline

\rowcolor{black!6}
\textbf{SGTD (Ours)}
& $\mathbf{77.30}$ & $\mathbf{91.17}$ & $\mathbf{48.34}$ & $\mathbf{72.27}$ & $\mathbf{28.96}$
& $\mathbf{67.18}$ & $\mathbf{89.97}$ & $\mathbf{35.93}$ & $\mathbf{64.36}$ & $\mathbf{31.25}$
& $\mathbf{79.11}$ & $\underline{91.92}$ & $\mathbf{51.32}$ & $\mathbf{74.11}$ & $\mathbf{27.79}$
\tabularnewline

\bottomrule
\end{tabularx}
\endgroup

\vspace{-5pt}
\end{table*}

\begin{table*}[t]
\centering

\caption{\textbf{Results on VisDiaHalBench (\%).} Avg. is the mean over Clean, Fact., and Hallu., while H.D. is Clean minus Hallu.}
\label{tab:visdiahalbench_paired_baselines}

\begingroup
\scriptsize
\setlength{\tabcolsep}{1.3pt}
\renewcommand{\arraystretch}{0.96}

\begin{tabularx}{\linewidth}{
    >{\raggedright\arraybackslash}p{1.55cm}
    *{5}{>{\centering\arraybackslash}X}
    @{\hspace{1.5pt}}
    !{\color{black!25}\vrule width 0.45pt}
    @{\hspace{1.5pt}}
    *{5}{>{\centering\arraybackslash}X}
    @{\hspace{1.5pt}}
    !{\color{black!25}\vrule width 0.45pt}
    @{\hspace{1.5pt}}
    *{5}{>{\centering\arraybackslash}X}
}

\toprule
\multirow{2}{*}{\textbf{Method}}
& \multicolumn{5}{c}{\textbf{LLaVA-1.5-7B}}
& \multicolumn{5}{c}{\textbf{mPLUG-Owl2-7B}}
& \multicolumn{5}{c}{\textbf{ShareGPT4V-7B}}
\tabularnewline

\cmidrule(lr){2-6}
\cmidrule(lr){7-11}
\cmidrule(lr){12-16}

& \mbox{Clean $\uparrow$}
& \mbox{Fact. $\uparrow$}
& \mbox{Hallu. $\uparrow$}
& \mbox{Avg. $\uparrow$}
& \mbox{H.D. $\downarrow$}
& \mbox{Clean $\uparrow$}
& \mbox{Fact. $\uparrow$}
& \mbox{Hallu. $\uparrow$}
& \mbox{Avg. $\uparrow$}
& \mbox{H.D. $\downarrow$}
& \mbox{Clean $\uparrow$}
& \mbox{Fact. $\uparrow$}
& \mbox{Hallu. $\uparrow$}
& \mbox{Avg. $\uparrow$}
& \mbox{H.D. $\downarrow$}
\tabularnewline

\midrule

Regular
& $77.16$ & $86.38$ & $30.72$ & $64.75$ & $46.44$
& $70.90$ & $86.10$ & $31.14$ & $62.71$ & $39.76$
& $79.04$ & $84.82$ & $23.18$ & $62.35$ & $55.86$
\tabularnewline

VCD~\cite{leng2024vcd}
& $77.60$ & $86.80$ & $32.04$ & $65.48$ & $45.56$
& $69.14$ & $86.64$ & $34.08$ & $63.29$ & $35.06$
& $79.86$ & $86.02$ & $24.50$ & $63.46$ & $55.36$
\tabularnewline

M3ID~\cite{favero2024m3id}
& $79.10$ & $\underline{87.32}$ & $30.74$ & $65.72$ & $48.36$
& $72.42$ & $\underline{87.18}$ & $31.20$ & $63.60$ & $41.22$
& $\underline{81.10}$ & $86.12$ & $23.48$ & $63.57$ & $57.62$
\tabularnewline

PAI~\cite{liu2024pai}
& $\underline{79.92}$ & $87.24$ & $31.34$ & $66.17$ & $48.58$
& $72.44$ & $87.08$ & $31.90$ & $63.81$ & $40.54$
& $80.68$ & $85.78$ & $24.02$ & $63.49$ & $56.66$
\tabularnewline

SID~\cite{huo2025sid}
& $78.72$ & $86.86$ & $33.84$ & $66.47$ & $44.88$
& $\underline{73.90}$ & $86.40$ & $30.82$ & $63.71$ & $43.08$
& $80.16$ & $85.74$ & $25.68$ & $63.86$ & $54.48$
\tabularnewline

AvisC~\cite{woo2025avisc}
& $79.18$ & $86.30$ & $21.24$ & $62.24$ & $57.94$
& $69.72$ & $83.96$ & $35.10$ & $62.93$ & $34.62$
& $79.94$ & $\underline{86.62}$ & $11.84$ & $59.47$ & $68.10$
\tabularnewline

CAI~\cite{lei2026cai}
& $70.16$ & $66.96$ & $21.94$ & $53.02$ & $48.22$
& $62.46$ & $82.30$ & $32.52$ & $59.09$ & $29.94$
& $73.22$ & $74.92$ & $18.54$ & $55.56$ & $54.68$
\tabularnewline

\midrule

RVD~\cite{zhong2024investigating}
& $77.16$ & $87.20$ & $\underline{43.36}$ & $\underline{69.24}$ & $\underline{33.80}$
& $70.90$ & $78.68$ & $\underline{59.70}$ & $\underline{69.76}$ & $\underline{11.20}$
& $79.04$ & $86.10$ & $\underline{37.18}$ & $\underline{67.44}$ & $\underline{41.86}$
\tabularnewline

\rowcolor{black!6}
\textbf{SGTD (Ours)}
& $\mathbf{83.06}$ & $\mathbf{91.28}$ & $\mathbf{83.04}$ & $\mathbf{85.79}$ & $\mathbf{0.02}$
& $\mathbf{77.18}$ & $\mathbf{88.12}$ & $\mathbf{75.80}$ & $\mathbf{80.37}$ & $\mathbf{1.38}$
& $\mathbf{83.96}$ & $\mathbf{89.74}$ & $\mathbf{80.60}$ & $\mathbf{84.77}$ & $\mathbf{3.36}$
\tabularnewline

\bottomrule
\end{tabularx}
\endgroup

\vspace{-5pt}
\end{table*}

\subsection{Datasets and Evaluation Metrics}
\label{sec:datasets_metrics}

We evaluate SGTD on the test sets of MMHalSnowball~\cite{zhong2024investigating} and VisDiaHalBench~\cite{cao2024visdiahalbench}.
MMHalSnowball contains 4,973 image and question pairs with factual and hallucinated histories constructed around each target question.
VisDiaHalBench contains 5,000 dialogues of five turns, where the model answers the final question using the original image and the preceding dialogue.
Both benchmarks use the three history conditions defined in Sec.~\ref{sec:problem_formulation}.
Clean provides no dialogue history, while Fact. and Hallu. provide histories with factual and hallucinated responses, respectively.
Following each benchmark's official answer-matching protocol, we compute accuracy (\%) under all three conditions.
Fact. accuracy measures performance with reliable context, whereas Hallu. accuracy measures performance when the history is misleading.
Avg. is the mean of the three accuracies and summarizes performance across history conditions.
H.D. is computed as Clean minus Hallu. and measures the degradation caused by hallucinated history.
Higher accuracy and Avg. are better, while a lower H.D. indicates stronger robustness.

\subsection{Implementation Details}
\label{sec:implementation_details}

We evaluate LLaVA-1.5-7B~\cite{liu2024improved}, mPLUG-Owl2-7B~\cite{ye2024mplugowl2}, and ShareGPT4V-7B~\cite{chen2024sharegpt4v}.
All LVLM parameters remain frozen.
Historical Answer Support Estimation uses frozen BLIP-VQA-base~\cite{li2022blip} to score each checkable historical answer against its type-matched alternatives.
We precompute these relative support scores once and use $\tau_{\mathrm{sup}}=0.5$ for Low-Support Selection.
In Support-Targeted Attention Intervention, we set the attention penalty to $\lambda=4$.
The Answer-Free Visual Branch uses a separate frozen BLIP-ITM-base-COCO~\cite{li2022blip} with a regional window ratio of $0.75$.
For Triadic Decoding, we set $(\alpha_{\mathrm{hist}},\omega)=(2,0.75)$ and $(\beta_{\min},\beta_{\max})=(0.05,0.5)$.
We use multinomial sampling with temperature 1.0 and conduct all experiments on NVIDIA A100 40GB GPUs.


\subsection{Main Results}
\label{sec:main_results}

\noindent\textbf{MMHalSnowball.}
Tab.~\ref{tab:mmhalsnowball_baselines} tests whether a method can resist hallucinated history without losing the benefit of factual context.
The single-turn baselines remain substantially below RVD under Hallu., showing that stronger visual grounding alone does not prevent the model from following misleading dialogue history.
SGTD ranks first in Hallu. and Avg. and obtains the lowest H.D. with all three LVLMs.
Compared with RVD, the strongest baseline under Hallu., SGTD improves Hallu. by 15.72\%, 13.79\%, and 17.48\%, while also improving Fact. by 7.30\%, 8.15\%, and 6.08\%.
In contrast, RVD reduces Fact. relative to Regular by 4.79\% to 5.87\% across the three backbones.
SGTD keeps Fact. 12.81\% to 22.79\% above Clean, showing that its robustness to misleading history does not weaken useful context.

\noindent\textbf{VisDiaHalBench.}
Tab.~\ref{tab:visdiahalbench_paired_baselines} shows that SGTD's advantage extends to five-turn dialogue.
RVD remains the strongest baseline under Hallu. across three LVLMs, ahead of the single-turn methods.
SGTD improves Hallu. over RVD by 39.68\%, 16.10\%, and 43.42\%, while raising Fact. by 4.08\%, 9.44\%, and 3.64\%.
Its H.D. is only 0.02\%, 1.38\%, and 3.36\%, nearly closing the gap between Clean and Hallu. on VisDiaHalBench.
The benchmark's five-turn structure helps explain this pattern.
A hallucinated history may contain useful questions and factual turns alongside misleading answers.
SGTD reduces attention to low-support spans while preserving supported context in the Support-Refined History Branch and previous questions in the Answer-Free Visual Branch.
With less interference from erroneous turns, the remaining history can guide the final query, allowing Hallu. to approach Clean.
By contrast, MMHalSnowball constructs each history around a single biased target, leaving fewer useful cross-turn cues after the misleading span is suppressed.
This difference explains why the gap between Clean and Hallu. is larger on MMHalSnowball.

\subsection{Ablation Studies}
\label{sec:ablation_studies}

\begin{table}[t]
    \centering
    \caption{\textbf{Ablation of Historical Answer Support Estimation.} Results are reported for LLaVA-1.5-7B (\%). RVS and LSS denote Relative Visual Support and Low-Support Selection.}
    \label{tab:history_support_ablation}
    \begingroup
    \normalsize
    \setlength{\tabcolsep}{0.8pt}
    \renewcommand{\arraystretch}{1.05}
    \resizebox{\columnwidth}{!}{%
    \begin{tabular}{cc@{\hspace{2pt}}ccc>{\columncolor{black!6}}c@{\hspace{2pt}}ccc>{\columncolor{black!6}}c}
        \toprule
        \multicolumn{2}{c}{\textbf{Setting}}
        & \multicolumn{4}{c}{\textbf{MMHalSnowball}}
        & \multicolumn{4}{c}{\textbf{VisDiaHalBench}} \\
        \cmidrule(lr){1-2}\cmidrule(lr){3-6}\cmidrule(lr){7-10}
        RVS & LSS
        & Clean $\uparrow$ & Fact. $\uparrow$ & Hallu. $\uparrow$ & Avg. $\uparrow$
        & Clean $\uparrow$ & Fact. $\uparrow$ & Hallu. $\uparrow$ & Avg. $\uparrow$ \\
        \midrule
        &
        & $\mathbf{77.30}$ & $\mathbf{91.88}$ & $24.33$ & $64.50$
        & $\mathbf{83.06}$ & $\underline{90.92}$ & $61.36$ & $78.45$ \\
        $\checkmark$ &
        & $\mathbf{77.30}$ & $79.25$ & $\mathbf{51.23}$ & $\underline{69.26}$
        & $\mathbf{83.06}$ & $89.66$ & $\underline{82.77}$ & $\underline{85.16}$ \\
        \midrule
        $\checkmark$ & $\checkmark$
        & $\mathbf{77.30}$ & $\underline{91.17}$ & $\underline{48.34}$ & $\mathbf{72.27}$
        & $\mathbf{83.06}$ & $\mathbf{91.28}$ & $\mathbf{83.04}$ & $\mathbf{85.79}$ \\
        \bottomrule
    \end{tabular}%
    }
    \endgroup
\end{table}

\noindent\textbf{Historical Answer Support Estimation.}
RVS scoring substantially improves Hallu. accuracy on both benchmarks.
Applying it to every checkable turn also reduces Fact. because supported history is unnecessarily altered.
Low-Support Selection restores Fact. to 91.17\% and 91.28\% while retaining Hallu. accuracy of 48.34\% and 83.04\%.
RVS therefore provides the grounding signal, while selective intervention protects reliable context.

\begin{table}[t]
    \centering
    \caption{\textbf{Ablation of Decoding Branches.} Results are reported for LLaVA-1.5-7B (\%). Full, Ref., and Vis. denote the Full-History, Support-Refined History, and Answer-Free Visual Branches.}
    \label{tab:branch_ablation}
    \begingroup
    \normalsize
    \setlength{\tabcolsep}{0.8pt}
    \renewcommand{\arraystretch}{1.05}
    \resizebox{\columnwidth}{!}{%
    \begin{tabular}{ccc@{\hspace{2pt}}ccc>{\columncolor{black!6}}c@{\hspace{2pt}}ccc>{\columncolor{black!6}}c}
        \toprule
        \multicolumn{3}{c}{\textbf{Branch}}
        & \multicolumn{4}{c}{\textbf{MMHalSnowball}}
        & \multicolumn{4}{c}{\textbf{VisDiaHalBench}} \\
        \cmidrule(lr){1-3}\cmidrule(lr){4-7}\cmidrule(lr){8-11}
        Full & Ref. & Vis.
        & Clean $\uparrow$ & Fact. $\uparrow$ & Hallu. $\uparrow$ & Avg. $\uparrow$
        & Clean $\uparrow$ & Fact. $\uparrow$ & Hallu. $\uparrow$ & Avg. $\uparrow$ \\
        \midrule
        & $\checkmark$ & $\checkmark$
        & $\mathbf{77.30}$ & $\underline{91.19}$ & $\underline{45.59}$ & $\underline{71.36}$
        & $\mathbf{83.06}$ & $\mathbf{91.68}$ & $\underline{81.60}$ & $\underline{85.45}$ \\
        $\checkmark$ & & $\checkmark$
        & $\mathbf{77.30}$ & $\mathbf{91.74}$ & $24.39$ & $64.47$
        & $\mathbf{83.06}$ & $90.86$ & $61.38$ & $78.43$ \\
        $\checkmark$ & $\checkmark$ &
        & $\underline{71.00}$ & $88.08$ & $39.05$ & $66.04$
        & $\underline{77.16}$ & $85.60$ & $72.66$ & $78.47$ \\
        \midrule
        $\checkmark$ & $\checkmark$ & $\checkmark$
        & $\mathbf{77.30}$ & $91.17$ & $\mathbf{48.34}$ & $\mathbf{72.27}$
        & $\mathbf{83.06}$ & $\underline{91.28}$ & $\mathbf{83.04}$ & $\mathbf{85.79}$ \\
        \bottomrule
    \end{tabular}%
    }
    \endgroup
\end{table}

\noindent\textbf{Decoding Branches.}
Removing the Support-Refined History Branch causes the largest Hallu. degradation, reducing accuracy by 23.95\% on MMHalSnowball and 21.66\% on VisDiaHalBench.
Selectively weakening low-support history provides the main defense against misleading context.
Removing the Answer-Free Visual Branch lowers performance across conditions and reduces Avg. by 6.23\% and 7.32\%, respectively.
Excluding the Full-History Branch lowers Hallu. by 2.75\% and 1.44\%, showing that its matched reference still strengthens the targeted correction.
The branches provide complementary refinement, visual grounding, and contextual reference signals.

\begin{figure}[t]
    \centering
    \includegraphics[width=\columnwidth]{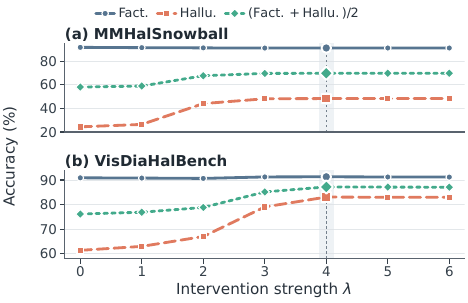}
    \caption{\textbf{Sensitivity to intervention strength.} Results are reported for (a) MMHalSnowball and (b) VisDiaHalBench with LLaVA-1.5-7B. Curves report Fact., Hallu., and $(\mathrm{Fact.}+\mathrm{Hallu.})/2$. The shaded marker denotes the selected $\lambda=4$ across both benchmarks. Setting $\lambda=0$ disables the attention penalty.}
    \label{fig:intervention_strength}
\end{figure}

\noindent\textbf{Intervention Strength.}
Fig.~\ref{fig:intervention_strength} examines penalty strength $\lambda$ for Support-Targeted Attention Intervention.
As $\lambda$ increases, Hallu. improves sharply on both benchmarks while Fact. remains stable, showing that stronger penalties suppress low-support history without weakening reliable context.
Their joint trend rises before reaching a broad plateau near the selected setting.
Further intervention produces marginal variation, indicating stability across strengths.
We therefore use $\lambda=4$ as the first stable operating point.

\begin{table}[t]
    \centering
    \caption{\textbf{Ablation of the Support-Adaptive Plausibility Constraint.} Results are reported for LLaVA-1.5-7B (\%).}
    \label{tab:plausibility_constraint_ablation}
    \begingroup
    \normalsize
    \setlength{\tabcolsep}{0.8pt}
    \renewcommand{\arraystretch}{1.05}
    \resizebox{\columnwidth}{!}{%
    \begin{tabular}{l@{\hspace{2pt}}ccc>{\columncolor{black!6}}c@{\hspace{2pt}}ccc>{\columncolor{black!6}}c}
        \toprule
        \multicolumn{1}{c}{\textbf{Constraint}}
        & \multicolumn{4}{c}{\textbf{MMHalSnowball}}
        & \multicolumn{4}{c}{\textbf{VisDiaHalBench}} \\
        \cmidrule(lr){1-1}\cmidrule(lr){2-5}\cmidrule(lr){6-9}
        & Clean $\uparrow$ & Fact. $\uparrow$ & Hallu. $\uparrow$ & Avg. $\uparrow$
        & Clean $\uparrow$ & Fact. $\uparrow$ & Hallu. $\uparrow$ & Avg. $\uparrow$ \\
        \midrule
        None
        & $75.21$ & $88.78$ & $\mathbf{49.24}$ & $71.08$
        & $81.96$ & $90.80$ & $\underline{82.36}$ & $\underline{85.04}$ \\
        Fixed $0.1$
        & $75.69$ & $89.93$ & $47.66$ & $\underline{71.09}$
        & $82.24$ & $\underline{91.34}$ & $81.54$ & $\underline{85.04}$ \\
        Fixed $0.2$
        & $75.93$ & $91.25$ & $44.46$ & $70.55$
        & $81.96$ & $\mathbf{91.38}$ & $79.26$ & $84.20$ \\
        Fixed $0.3$
        & $76.59$ & $91.55$ & $42.53$ & $70.23$
        & $82.94$ & $\underline{91.34}$ & $78.04$ & $84.11$ \\
        Fixed $0.4$
        & $77.26$ & $\underline{92.56}$ & $40.50$ & $70.11$
        & $\mathbf{83.30}$ & $90.96$ & $76.82$ & $83.69$ \\
        Fixed $0.5$
        & $\mathbf{77.70}$ & $\mathbf{92.94}$ & $39.35$ & $70.00$
        & $\underline{83.12}$ & $90.74$ & $75.92$ & $83.26$ \\
        \midrule
        \textbf{Adaptive}
        & $\underline{77.30}$ & $91.17$ & $\underline{48.34}$ & $\mathbf{72.27}$
        & $83.06$ & $91.28$ & $\mathbf{83.04}$ & $\mathbf{85.79}$ \\
        \bottomrule
    \end{tabular}%
    }
    \endgroup
\end{table}

\noindent\textbf{Support-Adaptive Plausibility Constraint.}
Tab.~\ref{tab:plausibility_constraint_ablation} compares removing the constraint, using a fixed $\beta$, and adapting $\beta_{t,k}$ from visual agreement and history support.
Increasing a fixed $\beta$ consistently harms Hallu., while Fact. rises on MMHalSnowball but peaks early on VisDiaHalBench.
By adapting the threshold to visual and historical support, the proposed constraint exceeds the best fixed setting in Avg. by 1.18\% and 0.75\% on the two benchmarks.

\begin{figure}[t]
    \centering
    \includegraphics[width=\columnwidth]{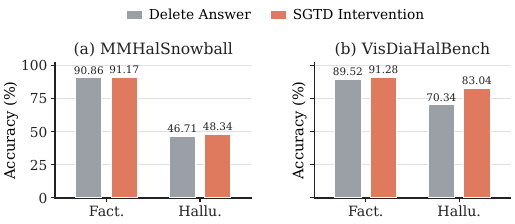}
    \caption{\textbf{Answer deletion versus attention intervention.} Accuracy on MMHalSnowball and VisDiaHalBench with LLaVA-1.5-7B (\%). Delete Answer removes the complete answer from each low-support turn, while SGTD keeps it in the history and applies Support-Targeted Attention Intervention to the selected span.}
    \label{fig:answer_deletion_ablation}
\end{figure}

\noindent\textbf{Answer Deletion or Intervention.}
Fig.~\ref{fig:answer_deletion_ablation} compares deleting the complete answer from each low-support turn with Support-Targeted Attention Intervention on LLaVA-1.5-7B, while retaining previous questions in both variants.
SGTD outperforms answer deletion under both history conditions, with a larger advantage on five-turn VisDiaHalBench.
Under Hallu., the improvement grows from 1.63\% on MMHalSnowball to 12.70\% on VisDiaHalBench.
Deletion discards reliable details, whereas targeted intervention preserves them as cross-turn dependencies accumulate.

\begin{table}[t]
    \centering
    \caption{\textbf{Generalization to Modern LVLM Backbones.} Accuracy on VisDiaHalBench (\%).}
    \label{tab:modern_backbone_ablation}
    \begingroup
    \small
    \setlength{\tabcolsep}{2.0pt}
    \renewcommand{\arraystretch}{1.02}
    \resizebox{\columnwidth}{!}{%
    \begin{tabular}{clccccc}
        \toprule
        \textbf{Backbone} & \textbf{Method}
        & Clean $\uparrow$ & Fact. $\uparrow$ & Hallu. $\uparrow$
        & Avg. $\uparrow$ & H.D. $\downarrow$ \\
        \midrule
        \multirow[c]{3}{*}{InternVL3.5-8B~\cite{wang2025internvl35}}
        & Regular & $\underline{75.58}$ & $\underline{87.66}$ & $36.38$ & $66.54$ & $39.20$ \\
        & RVD & $\underline{75.58}$ & $80.50$ & $\underline{72.44}$ & $\underline{76.17}$ & $\underline{3.14}$ \\
        & \cellcolor{black!6}\textbf{SGTD (Ours)}
        & \cellcolor{black!6}$\mathbf{78.84}$
        & \cellcolor{black!6}$\mathbf{90.56}$
        & \cellcolor{black!6}$\mathbf{82.78}$
        & \cellcolor{black!6}$\mathbf{84.06}$
        & \cellcolor{black!6}$\mathbf{-3.94}$ \\
        \midrule
        \multirow[c]{3}{*}{Qwen2.5-VL-7B~\cite{bai2025qwen25vl}}
        & Regular & $\underline{73.52}$ & $\mathbf{79.20}$ & $65.20$ & $72.64$ & $8.32$ \\
        & RVD & $\underline{73.52}$ & $75.28$ & $\underline{71.18}$ & $\underline{73.33}$ & $\underline{2.34}$ \\
        & \cellcolor{black!6}\textbf{SGTD (Ours)}
        & \cellcolor{black!6}$\mathbf{75.20}$
        & \cellcolor{black!6}$\underline{79.18}$
        & \cellcolor{black!6}$\mathbf{78.00}$
        & \cellcolor{black!6}$\mathbf{77.46}$
        & \cellcolor{black!6}$\mathbf{-2.80}$ \\
        \midrule
        \multirow[c]{3}{*}{Qwen3-VL-8B~\cite{bai2025qwen3vl}}
        & Regular & $\underline{80.86}$ & $\mathbf{88.44}$ & $54.62$ & $74.64$ & $26.24$ \\
        & RVD & $\underline{80.86}$ & $82.04$ & $\underline{76.90}$ & $\underline{79.93}$ & $\underline{3.96}$ \\
        & \cellcolor{black!6}\textbf{SGTD (Ours)}
        & \cellcolor{black!6}$\mathbf{81.52}$
        & \cellcolor{black!6}$\underline{87.10}$
        & \cellcolor{black!6}$\mathbf{80.66}$
        & \cellcolor{black!6}$\mathbf{83.09}$
        & \cellcolor{black!6}$\mathbf{0.86}$ \\
        \midrule
        \multirow[c]{3}{*}{LLaVA-NeXT-1.6~\cite{liu2024llavanext}}
        & Regular & $\underline{69.76}$ & $\underline{87.50}$ & $38.18$ & $65.15$ & $31.58$ \\
        & RVD & $\underline{69.76}$ & $85.10$ & $\underline{51.34}$ & $\underline{68.73}$ & $\underline{18.42}$ \\
        & \cellcolor{black!6}\textbf{SGTD (Ours)}
        & \cellcolor{black!6}$\mathbf{76.06}$
        & \cellcolor{black!6}$\mathbf{89.90}$
        & \cellcolor{black!6}$\mathbf{84.66}$
        & \cellcolor{black!6}$\mathbf{83.54}$
        & \cellcolor{black!6}$\mathbf{-8.60}$ \\
        \bottomrule
    \end{tabular}%
    }
    \endgroup
\end{table}

\noindent\textbf{Modern LVLM Backbones.}
Tab.~\ref{tab:modern_backbone_ablation} examines whether SGTD transfers to four recent LVLM backbones on VisDiaHalBench.
SGTD achieves the highest Hallu. and Avg. accuracy for every backbone, improving Hallu. over RVD by 3.76\% to 33.32\%.
Its H.D. is also the lowest throughout and becomes negative on three backbones.
This occurs because SGTD suppresses misleading turns while retaining useful cross-turn cues unavailable in Clean, allowing Hallu. accuracy to exceed Clean.

\subsection{Qualitative Analysis}
\label{sec:qualitative_analysis}

\begin{figure}[t]
    \centering
    \includegraphics[width=\columnwidth]{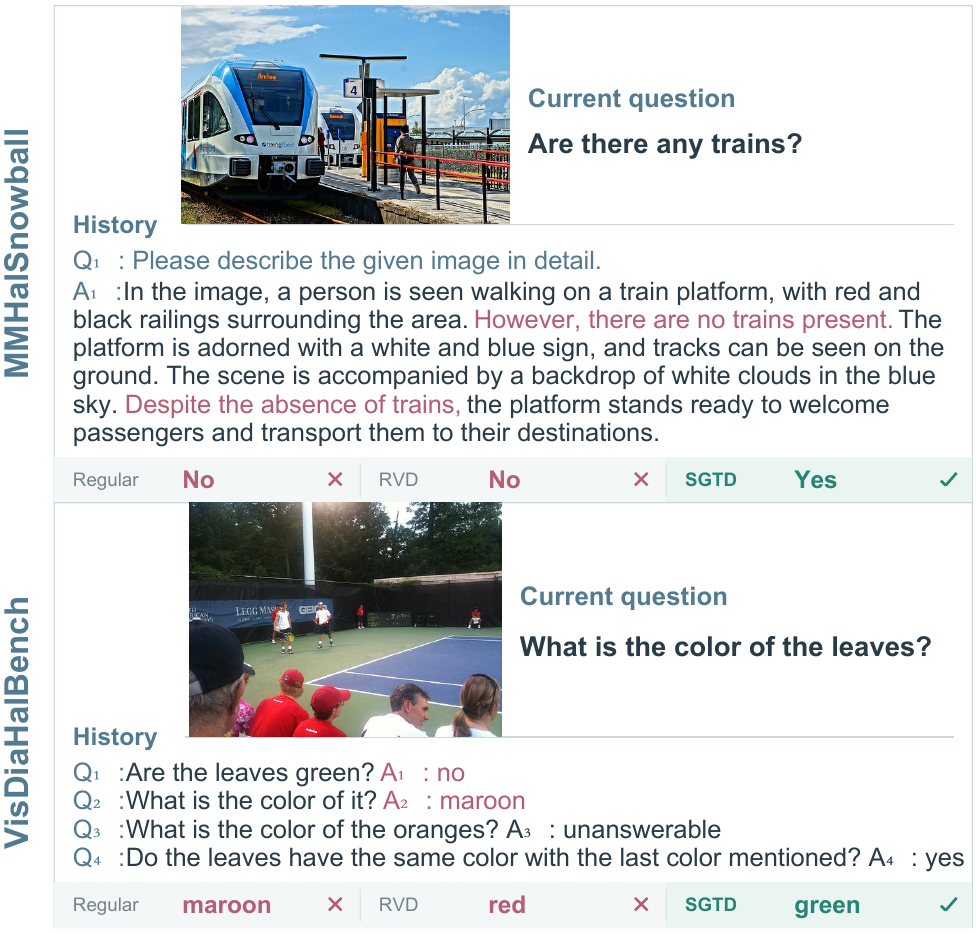}
    \caption{\textbf{Qualitative comparison under hallucinated histories.} The top and bottom rows show examples from MMHalSnowball and VisDiaHalBench, respectively, using LLaVA-1.5-7B. Regular decoding and RVD follow unsupported history, whereas SGTD recovers the image-supported answers. Red and green denote incorrect and correct predictions.}
    \label{fig:qualitative_comparison}
\end{figure}

Fig.~\ref{fig:qualitative_comparison} shows how unsupported history affects dialogues of different lengths.
On MMHalSnowball, the history falsely denies the presence of the visible train, leading Regular decoding and RVD to answer ``No.''
On VisDiaHalBench, later turns reinforce the false color history, leading Regular decoding and RVD to predict ``maroon'' and ``red.''
SGTD identifies the low-support content and combines the Support-Refined History and Answer-Free Visual Branches to recover ``Yes'' and ``green,'' while preserving useful context and correcting propagated errors.

\section{Conclusion}
\label{sec:conclusion}

Multimodal dialogue requires controlling misleading history without losing useful cross-turn context.
We introduced Support-Guided Triadic Decoding (SGTD), a training-free framework that estimates historical answer support and coordinates predictions from the Full-History, Support-Refined History, and Answer-Free Visual Branches.
Across two benchmarks and representative LVLMs, SGTD consistently improves robustness to hallucinated histories while retaining gains from factual context, demonstrating the value of support-guided history refinement and Triadic Decoding.


\section*{LLM Usage Statement}
OpenAI Codex powered by GPT-5.6 Sol was employed to support language polishing, manuscript structuring, and improvements in writing style. The model also helped refine technical explanations throughout the main paper and supplementary materials to enhance clarity and consistency. All methodological development, experimental design and execution, result analysis, and scientific conclusions were independently carried out by the authors. The authors carefully reviewed and verified all AI-assisted content, including cross-checking numerical claims against the original experimental results. Full responsibility for the research methodology, findings, interpretations, and final manuscript rests with the authors.

{
    \small
    \bibliographystyle{ieeenat_fullname}
    \bibliography{main}
}


\end{document}